%% file: main.tex
\documentclass[10pt,twocolumn,letterpaper]{article}
\usepackage[accsupp]{axessibility}  
\usepackage{iccv}
\usepackage{times}
\usepackage{epsfig}
\usepackage{graphicx}
\usepackage[numbers,sort]{natbib} 
\usepackage{amsmath}
\usepackage{amssymb}
\usepackage{soul}
\usepackage{booktabs}
\usepackage{caption}
\usepackage{scalerel}
\usepackage{comment}
\usepackage{caption}
\usepackage{url}
\usepackage{hyperref}
\hypersetup{breaklinks=true,bookmarks=false}

\iccvfinalcopy 

\def\iccvPaperID{****} 

\usepackage[capitalize]{cleveref}
\crefname{section}{Sec.}{Secs.}
\Crefname{section}{Section}{Sections}
\Crefname{table}{Table}{Tables}
\crefname{table}{Tab.}{Tabs.}

\usepackage{xcolor}         
\newcommand{\modified}[1]{\textcolor{black}{#1}}

\newcommand{\weidi}[1]{\textcolor{red}{[Weidi: #1]}}

\newcommand*{\score}{\ensuremath{S_{\alpha}}}
\newcommand*{\corrected}[1]{\textcolor{black}{#1}}
\newcommand*{\cor}[1]{\textcolor{black}{#1}}
\def\iccvPaperID{5370} 

\ificcvfinal\fi

\begin{document}

\title{The Making and Breaking of Camouflage}

\author{Hala Lamdouar$^1$ \qquad  \text{Weidi Xie}$^{1,2}$ \qquad  \text{Andrew Zisserman}$^1$\\
$^1$ \text{Visual Geometry Group, University of Oxford}  \qquad $^2$ \text{CMIC, Shanghai Jiao Tong University}\\
{\tt\small \{lamdouar,weidi,az\}@robots.ox.ac.uk}}


\maketitle


\input{text/00-abstract.tex}

\input{text/01-introduction.tex}
\input{text/02-related_work.tex}

\input{text/03-measuring_camouflage.tex}
\input{text/04-pipeline.tex}

\input{text/05-experiments.tex}
\input{text/06-conclusion.tex}

{\small
\bibliographystyle{ieee_fullname}
\bibliography{all_bib}
}

\end{document}

%% file: text/00-abstract.tex
\begin{abstract}
   Not all camouflages are equally effective, 
   as even a partially visible contour or a slight color difference can make the animal stand out and break its camouflage. 
   In this paper, we address the question of what makes a camouflage successful, by proposing three scores for automatically assessing its effectiveness.
   In particular, we show that camouflage can be measured by the similarity between background and foreground features and boundary visibility. We use these camouflage scores to assess and compare {\em all} available camouflage datasets.
   We also incorporate the proposed camouflage score into a generative model as an auxiliary loss and show that effective camouflage images or videos can be synthesised in a scalable manner. The  generated synthetic dataset is used to train a transformer-based model for segmenting camouflaged animals in videos. Experimentally, we demonstrate state-of-the-art camouflage breaking performance on the public MoCA-Mask benchmark. 
\end{abstract}

%% file: text/01-introduction.tex
\section{Introduction}
\label{sec:intro}
Camouflage has long been a subject of interest and fascination for the scientific community, especially evolutionary \corrected{biologists} who consider it as an excellent example of species adaptation. In order to confuse \corrected{predators or to hide from prey},
and increase their chances of survival in their natural habitat, animals have developed numerous camouflage mechanisms, {\em e.g.}, disruptive coloration and background matching. Some species have even evolved to develop an adaptive camouflage, {\em e.g.}, an arctic fox loses its white fur to better match the brown grey of the new season’s landscape. Perhaps the most dramatic camouflage adaptation is the cuttlefish; it changes its patterns dynamically and rapidly as it moves from one spot to the other, constantly adapting and improving its camouflage. 

\begin{figure}[t]
   \includegraphics[width=\columnwidth]{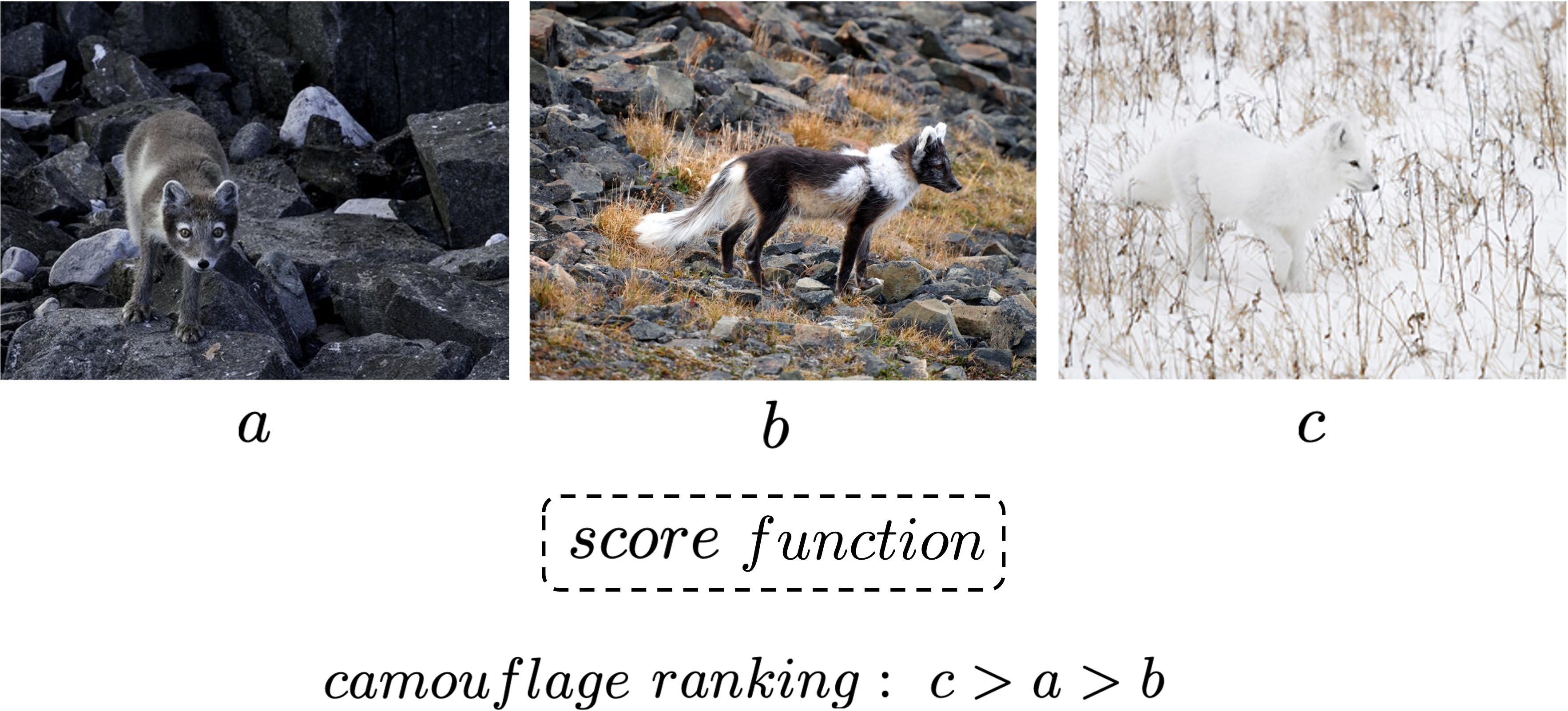}
   \caption{The three images depict the same animal, an Arctic fox, as it adapts its appearance to better blend with the changing landscape of the new season. While images $a$ and $c$ exhibit better background similarity than image $b$, the fox boundary is more visible in image $a$ than image $c$. We assess the effectiveness of camouflage by measuring the degree of ambiguity it creates with respect to its background.}
   \label{fig:teaser_1}
   \vspace{-0.5cm}
\end{figure}

This search for optimal camouflage has inspired numerous works in the computer vision community, such as~\cite{Owens14, Guo22}, 
that tackled camouflage as a problem of optimal texture synthesis, \cor{making} 3D objects non-detectable in a given scene. Others have addressed camouflage as a highly challenging object segmentation task in~\cite{Fan21,Le19,Mei21,kajiura2021improving}. 
Efforts have been made in collecting large scale camouflage datasets~\cite{Fan21,Le19} with costly annotation. 
In fact, camouflaged animals often exhibit complex shapes and thin structures that add to the boundary ambiguity and make the annotation highly time-consuming. 
Fan {\em et al.\ }report up to $60$min per image to provide accurate pixel-level annotations for their dataset COD10K~\cite{Fan21}.
While another line of research turned to camouflage breaking in sequences by taking advantage of motion cues~\cite{Lamdouar20,Bideau16,Lamdouar21,Yang21a,Cheng22}, 
the camouflage data scarcity is even more extreme for videos. 
Recently, Sim2Real~\cite{Lamdouar21,xie22} training has shown to be very effective for motion segmentation. \cor{By} training on the optical flows of synthetic videos, \cor{these models} can generalise to real videos without suffering from the domain gaps.

In this paper, we start by asking the question:
``\textbf{what are the properties of \cor{a successful} camouflage ?}''
To answer this question, we investigate three \cor{scoring functions} for quantifying the effectiveness of camouflage, namely, reconstruction fidelity score ($S_{R_f}$), boundary score ($S_b$) and intra-image Fr\'echet score ($d_{\mathcal{F}}$). 
These scores are later adopted for two critical roles,
(i) they are used to assess the relevance of existing camouflage datasets and act as quality-indicator in collecting new camouflage data;
(ii) they can be used as a proxy loss for image and video inpainting/generation, 
where we establish a synthetic camouflage data generation pipeline with a \corrected{Generative} Adversarial Network~(GAN), that can simultaneously generate high-quality camouflage \corrected{examples and masks} of the camouflaged animals. 
We further train \corrected{a} Transformer-based architecture on the generated camouflage video sequences, and demonstrate \corrected{a} performance boost over training on only the (small scale) real data.





In summary, we make the following contributions:
{\em First}, we introduce three scoring functions to measure the effectiveness of a given camouflage in still images and videos.
We use these  camouflage scores to rank the images in existing datasets in terms of camouflage success,  and also show that the rankings are largely in accordance with human-produced rankings. 
{\em Second}, we incorporate the camouflage score into a generative model,
establishing a scalable pipeline for generating high-quality camouflage images or videos, along with the pixelwise segmentation mask of the camouflaged animals;
{\em Third}, we show that a Transformer-based model trained on the synthetically generated data can achieve state-of-the-art performance on the \cor{MoCA-Mask} video camouflaged animal segmentation benchmark. 

%% file: text/02-related_work.tex
\section{Related work}

\label{sec:related_work}

\noindent \textbf{Camouflage Evaluation.}
Although there are no previous computational works that directly assess camouflage, as far as we know, there have been a number of human studies.
Previous works proposed methods to evaluate camouflage by analysing the human viewers' eye movements~\cite{Lin14,troscianko17} or conducting subjective perceptual experiments~\cite{Owens14, Guo22, Skurowski18}. In~\cite{Owens14, Guo22}, participants were asked to indicate the camouflaged object and their answers were analysed in terms of accuracy and time needed per image. Similarly, Skurowski {\em et al.}~\cite{Skurowski18} asked human volunteers to rate CHAMELEON~\cite{Skurowski18} images from 1 to 5 in terms of camouflage effectiveness. They produced a score after compensating for personal bias. We validate our proposed camouflage scoring functions by comparing our rankings to these human-based rankings.

\vspace{3pt}
\noindent \textbf{Motion Segmentation.}
The goal of this task is to partition the frames of a video sequence into background and independently moving objects. Early approaches focused on clustering motion patterns by grouping pixels with similar trajectories~\cite{Brox10,Ochs11,Keuper15,Sivic04a,lezama11,ochs12,ochs13}. Another line of research tackled the problem by compensating for the background motion~\cite{Bideau16,Bideau18,Lamdouar20}, 
via registering consecutive frames, or explicitly estimating camera motion. More recently, in~\cite{Yang21a}, a Transformer-like architecture is used to reconstruct the input flows, and the segmentation masks \corrected{are} generated as a side product, while \cite{Choudhury22} decomposes the flow filed into regions by fitting affine transformations.
The \cor{most} closely related to ours is~\cite{Lamdouar21,xie22},
where the authors adopt a Sim2Real strategy 
by training the model on optical flow computed from synthetic videos. \cor{These} models can generalise towards real videos without fine-tuning.
In this paper, we  train a model on synthetic camouflage {\em RGB} sequences.



\vspace{3pt}
\noindent \textbf{Fr\'echet Inception Distance and its variants.} 
Also known as Wasserstein-2 distance~\cite{villani09}, 
Fr\'echet distance (FD) is a metric quantifying the difference between two probability distributions~\cite{frechet57}. 
Lately, \cite{Heusel17} introduced the Fr\'echet Inception Distance (FID) in the context of generative models. Under the assumption that real images and generated images follow Gaussian distributions, they compute the FD of the two distributions. Note that here, each image is represented by a vector obtained from the last pooling layer of InceptionV3~\cite{szegedy16}. 
Following their steps, \cite{Shaham19} adapted FID to the image re-targeting task and introduced the Single Image Fr\'echet Inception Distance (SIFID). Instead of comparing entire image datasets, they compare two images and consider the distributions of their feature vectors obtained from earlier layer of InceptionV3~\cite{szegedy16}. Recently, \cite{Doan20} included FID in the training objective and introduced Fr\'echet-GAN. Our work takes inspiration from~\cite{Shaham19,Doan20} and investigates a FID of regions within the same image as an auxiliary loss.  


%

%% file: text/03-measuring_camouflage.tex
\begin{figure*}[t]
\centering
\vspace{2pt}
\begin{minipage}[htb]{0.49\linewidth}
\centering
\includegraphics[width=\textwidth,height=9em]{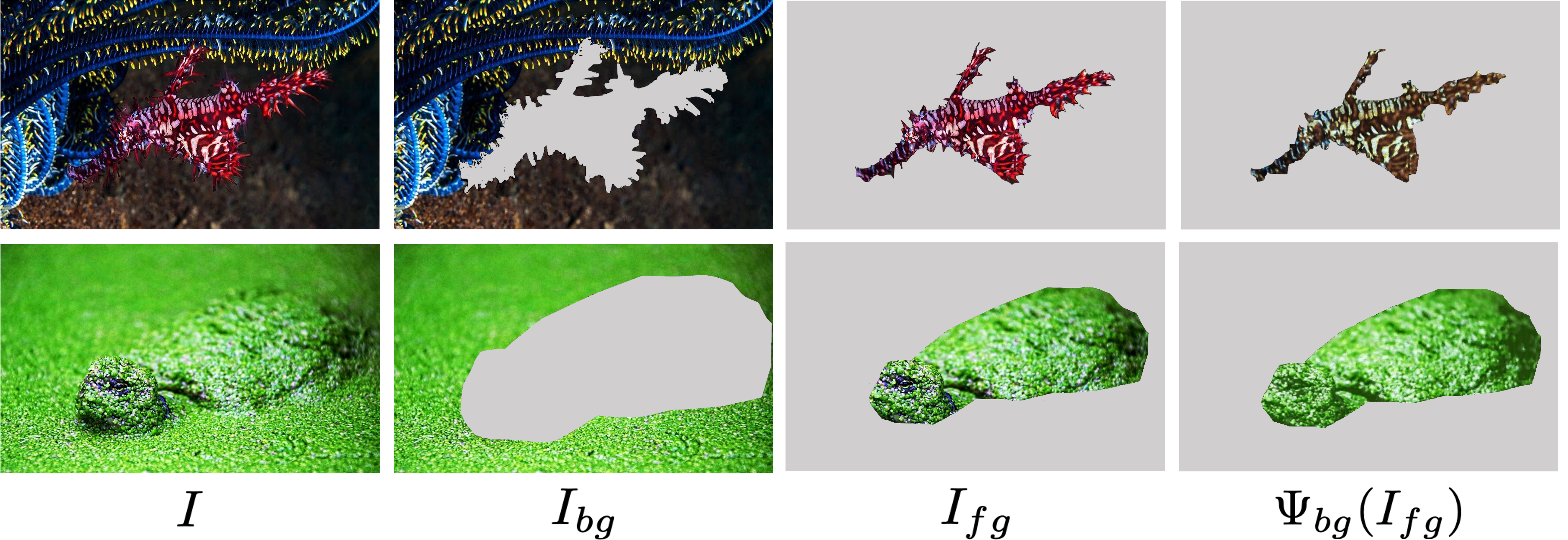}
\vspace{-0.5cm}
\caption{ The reconstruction fidelity score $S_{R_f}$ evaluates the similarities between the original foreground $I_{fg}$ and its reconstruction from background features $\Psi_{bg}(I_{fg})$.
The top example shows a case where the animal exhibits different visual patterns (color) from its background ($S_{R_f}$=0.11), while the bottom example shows a better background matching ($S_{R_f}$=0.82).}
\label{fig:reconstruction}
\end{minipage}
\hfill
\begin{minipage}[h]{0.48\linewidth}
\centering
\includegraphics[width=\textwidth]{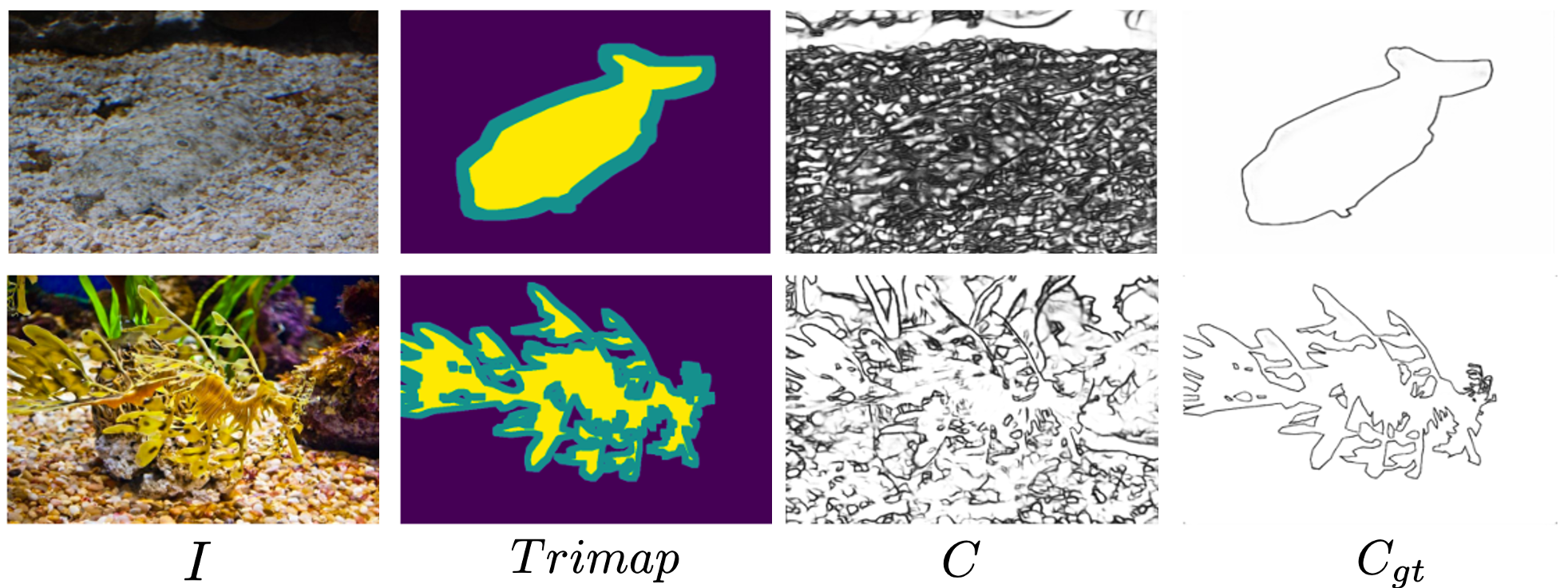}
\caption{Trimap regions and animal contours. From left to right: Input image; Trimap partition (foreground region in yellow, background in purple, and boundary region in green); Contours $C$ computed on the original image; and \modified{ground truth} Contours $C_{gt}$. The top example shows less contour agreement along the boundary region ($S_b$=0.72) than the bottom example ($S_b$=0.40).}
\label{fig:contour}
\end{minipage}
\vspace{-0.3cm}
\end{figure*}

\section{Measuring the effectiveness of camouflage}

Assuming there exists a dataset, 
$\mathcal{D} = \{(I_1, m_1), \dots, \allowbreak (I_N, m_N)\}$, 
where $I_i \in \mathbb{R}^{H \times W \times 3}$ refers to the images, and $m_i \in \{0,1\}^{H \times W \times 1}$ denotes a binary segmentation mask of the camouflaged animal. \cor{Here}, we aim to design a scoring function that takes the image and segmentation mask as input, and \corrected{outputs} a scalar value $S$ that can quantify the effectiveness of camouflage, {\em i.e.}, how successfully an animal blends into its background, termed as the camouflage score,
\begin{align*}
S : (I_i,m_i)&\longmapsto s_i \in [0,1]
\end{align*}
specifically, for $i\neq j$ and $s_i<s_j$, 
the animal in $I_j$, indicated by the mask $m_j$, is more concealed than the animal in $I_i$, indicated by the mask $m_i$. 
Having such a scoring function enables various applications,
{\em first}, it enables to rank the images from one dataset, 
in terms of the effectiveness of camouflage, 
and generate camouflage-relevant statistics for the dataset;
{\em second}, it can serve as the objective for low-level image generation tasks, for example, image inpainting, editing, etc.

In the following sections, we investigate \cor{three} such scoring functions,
specifically, we exploit the animal's mask and quantify the key aspects that determine the camouflage success both perceptually~(Sec.~\ref{sec:perceptual}) and probabilistically~(Sec.~\ref{sec:probabilistic}). Note that, we take into account the local aspect of camouflage, 
{\em i.e.}, instead of processing the entire background region, 
we only focus on the immediate surround of the animal. 
Hence, when referring to background, we mean the local background and only consider the cropped images centred around the object.

\subsection{Perceptual camouflage score}
\label{sec:perceptual}
In this section, we define the camouflage score through a perceptual lens, 
specifically, we measure the foreground and background similarity~(Sect.~\ref{sec:reconstruct}), along with the boundary visibility~(Sect.~\ref{sec:bg_vis}). 

\vspace{-5pt}
\subsubsection{Reconstruction fidelity score}
\label{sec:reconstruct}
We attempt to reconstruct the foreground region with patches from the background, and propose a score to quantitatively measure the discrepancy between the original image and its reconstruction. Intuitively, for successful camouflage, we should be able to reconstruct the foreground perfectly by copying patches from its close neighborhood.

Formally, for a given image~($I$) and segmentation mask~($m$), 
subscript has been ignored for simplicity, 
we consider a trimap separation and define the foreground and background regions using morphological erosion and dilation of the mask, 
\begin{align}
    I_{\text{fg}}, \text{ } I_{\text{bg}} = m_{\text{fg}} \odot I, \text{ } m_{\text{bg}} \odot I
\end{align}
where $m_{\text{fg}} = \text{erode}(m)$ and $m_{\text{bg}} = 1 - \text{dilate}(m)$.

To reconstruct the foreground region, 
we replace the foreground patch with the closest one in the background. 
Here a patch is a $n \times n$ pixel region ($n = 7$ in our case with an \corrected{overlap} of 3) 
and the  the patchwise similarity is computed by exploiting the low-level perceptual similarity, {\em i.e.}, comparing corresponding RGB values.
This reconstruction method is inspired by the texture generation and inpainting approaches, such as~\cite{Efros99}. 
In practice, we implement it with fast approximate nearest neighbor search for efficiency. 
 
The reconstruction fidelity score is computed by \cor{assessing} the difference value between the foreground region and its reconstruction, 
specifically, we count the number of foreground pixels that have been successfully reconstructed from the background:
\begin{align}
&S_{R_f}(I,m) = \frac{1}{N_{\text{fg}}}\sum_{(i,j) \in I_{\text{fg}}} R_f(i,j) \\
&R_f(i,j) = \begin{cases} 1 & \mbox{if } || I_{\text{fg}} - \Psi_{I_{\text{bg}}}(I_{\text{fg}}) ||_2 < \lambda || I_{\text{fg}} ||_2 \\ 0 & \mbox{otherwise} \end{cases}
\end{align}
$\Psi_{I_{bg}}(.)$  denotes the reconstruction operation, $N_{\text{fg}}=|m_{\text{fg}}|$ is the total number of pixels in the foreground region, and $\lambda$ is a threshold ($\lambda$=0.2 in our case). 
A higher $S_{R_f}$ score, means that the animal's visual attributes are well represented in the background. Conversely, a low $S_{R_f}$ indicates that the animal's appearance is composed of unique features that makes it standout from its surrounding. In Fig.~\ref{fig:reconstruction}, we present examples of camouflage evaluation by reconstruction fidelity. 


\vspace{-6pt}
\subsubsection{Boundary visibility score} 
\label{sec:bg_vis}
With the trimap representation, we can easily extract the boundary region, 
$I_{\text{b}} = m_{\text{b}} \odot I$, 
where $m_{\text{b}} = (1-m_{\text{bg}})-m_{\text{fg}}$. 
Here, we aim to measure the animal's boundary properties, {\em e.g.}, contour visibility, as they can also provide visually evident cues for camouflage breaking. In particular, we adopt an off-the-shelf contour extraction method~\cite{Poma20}, and run it on \cor{the} original image and foreground mask,
to generate the \cor{images'} contour~($C$) and \modified{ground truth} animal's contour~($C_\text{gt}$), as shown in Fig.~\ref{fig:contour}.

We express the agreement between predicted contours $C$ and ground truth contours $C_\text{gt}$ over the boundary region with \cor{the} F1 metric:
\begin{align}
S_{b}(I,m) = 1 - \text{F1}(m_{b}\odot C_{\text{gt}}, \text{ } m_{b}\odot C)
\end{align}
This score penalises the boundary pixels that are predicted as contour in both $C$ and $C_{\text{gt}}$. 
While we do not expect a perfect contour agreement for visible edges, 
we consider $S_{b}$ as a reasonable approximation, given the shape and size of the boundary region as a thin margin centred around the animal.
If a boundary pixel is predicted as a contour in $C_{\text{gt}}$ but not in $C$, this means that the animal's edge is not visible in the original image.
If a pixel is predicted as a contour in $C$ but not $C_{\text{gt}}$, 
this indicates the presence of distracting elements, such as complex texture patterns on the foreground animal, which also affects the visibility of the animal's contour and therefore improves its camouflage. A perfectly camouflaged animal would have little contour agreement and $S_{b} = 1$.


To get the final perceptual camouflage score, 
we linearly combine both reconstruction fidelity score and boundary visibility score:
\begin{align}
\score = (1-\alpha) S_{R_f} + \alpha S_{b}
\label{eq:tri-score}
\end{align}
We will describe our method for setting the weighting parameter $\alpha$ in the Experiment section \cor{(Sec.~\ref{sec:experiment})}.


\subsection{Probabilistic scoring function}
\label{sec:probabilistic}

In addition to using the low-level RGB representation,
in this section, we consider the pixelwise representation in the feature space, and propose a differentiable metric that compares the \cor{probabilistic} distribution between the foreground and background regions, 
which acts as a proxy for the score $\score $ described in previous section,
and can be used directly as a differentiable loss function in image generation. 


We consider the \textbf{Intra-Image Fr\'echet Inception Distance}, 
specifically,  we compute the feature maps for the foreground and background image regions: 
\begin{align}
    f_{fg}, \text{ } f_{bg} = \Phi_{\text{v}}(I_{fg}), \text{ } \Phi_{\text{v}}(I_{bg})
\end{align}
Here $\Phi_{\text{v}}(\cdot)$ denotes a pre-trained Inception network~\cite{szegedy15,szegedy16}, truncated at early layers. 
We take inspiration from~\cite{Shaham19}, and adapt the single image Fr\'echet distance to measure the deviation between feature distributions of different regions within the same image. We adopt the Fr\'echet hypothesis with respect to our regions, 
{\em i.e.}, the features of the foreground and background follow  multivariate Gaussian distributions: 
$f_{\text{fg}} \sim \mathcal{N}(\mu_{\text{fg}},\Sigma_{\text{fg}})$ and $f_{\text{bg}} \sim \mathcal{N}(\mu_{\text{bg}},\Sigma_{\text{bg}})$. 
The intra-image Fr\'echet distance can be formulated as follows:
\begin{align*}
\corrected{
d_{\mathcal{F}}^2(I,m) = ||\mu_{\text{fg}} - \mu_{\text{bg}}||_2^2 + Tr(\Sigma_{\text{fg}}+\Sigma_{\text{bg}}-2(\Sigma_{\text{fg}} \Sigma_{\text{bg}})^{1/2})}
\end{align*}
\textbf{Note that}, when $(I,m)$ is the output from a generative model, $d_{\mathcal{F}}^2$ is differentiable with respect to its parameters, 
and can therefore be used as an auxiliary loss for optimising the image generation procedure, {\em i.e.}, generate effective camouflage examples.


%% file: text/04-pipeline.tex
\begin{figure*}[t]
\centering
\vspace{2pt}
\begin{minipage}[htb]{\linewidth}
\centering
\includegraphics[width=0.90\textwidth]{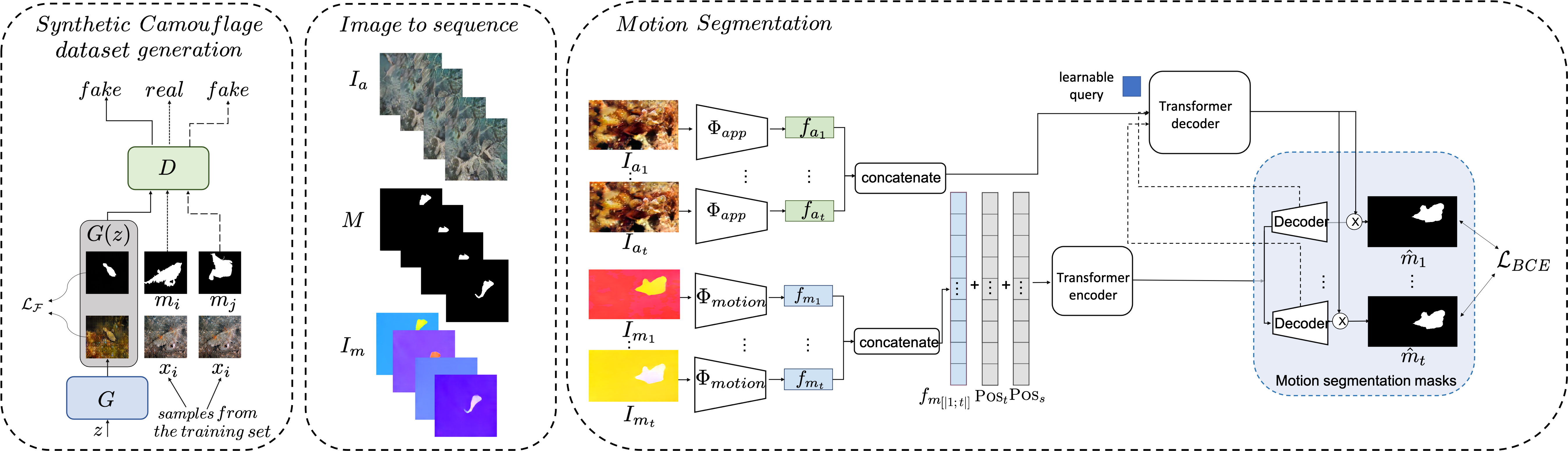}
\caption{Our framework consists of a synthetic camouflage data generation pipeline \cor{(left),} where we train a generator $G$, in a GAN setting, to create camouflage images and masks, while encouraging camouflage effectiveness via minimising $\mathcal{L}_{\mathcal{F}}$. The generated samples are then transformed into synthetic video sequences \cor{(middle),} following our method presented in Sec.~\ref{subsec:seq}. The transformer-based motion segmentation architecture \cor{(right),} for camouflage breaking in videos, is trained on the synthetic video sequences. The architecture is described in Sec.~\ref{sec:architecture}.}
\vspace{-0.5cm}
\label{fig:data}
\end{minipage}
\end{figure*}

\section{Generating camouflage video sequences}
In this section, we propose a scalable pipeline for generating images with concealed animals.
In particular, we incorporate the differentiable intra-image Fr\'echet distance in image generation, explicitly optimising the camouflage score of the image. We first detail the static image generation framework, 
then describe how we use the images it generates to create camouflage video sequences.

\subsection{Camouflage image generation}
\label{subsec:gan}
In order to generate the image with a camouflaged animal, 
and its corresponding segmentation mask, 
we adopt a Generative Adversarial Network~(GAN), 
where a generator is fed with a latent vector $z \sim \mathcal{N}(0,1)$ and learns to predict the pair of realistic image and segmentation mask, such that a discriminator cannot differentiate it from real samples.

Let $(x_i,m_i)$ denote an image and segmentation pair sampled from the training dataset, a subset of COD10K~\cite{Fan21} is used here.
We adopt the conventional label notation, 
{\em i.e.} $y=1$ for (real) examples sampled from the training dataset and $y=0$ for (fake) examples from the generator, \cor{and consider} the generative adversarial loss functions~\cite{Goodfellow14a}:
\begin{align*}
    \mathcal{L}_D &=  E_{\scaleto{(x_i,m_i)}{0.6em}}[-\log(D(x_i,m_i))]+E_{z}[-\log(1-D(G(z)))]\\
    \mathcal{L}_G &=  E_{z}[-\log(D(G(z)))]
\end{align*}

To enforce coherence between images and their segmentation masks, 
we feed the discriminator with additional fake pairs, 
consisting of real images coupled with unpaired real masks. 
This can be formulated as an additional coherence loss term that minimises the probability of assigning incorrect labels, {\em i.e.} $y=1$, to $(x_i,m_j)$ sampled from the training set such that $i \neq j$:
\begin{align}
\mathcal{L}_{\rm Coh} = E_{\scaleto{(x_i,m_j)_{i \neq j}}{0.6em}}[-\log(1-D(x_i,m_j))]
\end{align}
To increase the camouflage effectiveness in the generated examples, 
we adopt the intra-image Fr\'echet distance in the generator loss, 
as an auxiliary loss term, and train our camouflage image generator with the following loss:
\begin{align}
    \mathcal{L}_{\tilde{D}} =  \mathcal{L}_D + \mathcal{L}_{\rm Coh} ~~~~~~
    \mathcal{L}_{\tilde{G}} =  \mathcal{L}_G + \beta  d^2_{\mathcal{F}}
\end{align} 
We present an overview of our data generation framework in Fig.~\ref{fig:data}.


\subsection{Camouflage video generation}
\label{subsec:seq}
Given a camouflage image and corresponding segmentation mask, generated using the method above,
we can create video sequences by applying different motions to the foreground and background, in a similar manner to~\cite{Lamdouar21}.
Specifically, we first inpaint the backgrounds with an off-the-shelf model proposed by Suvorov {\em et al.}~\cite{Suvorov21} and overlay the synthetic animal (extracted from the original generated image by using the mask) at a random location within the image in the first frame, then following a random translational trajectory in the following frames. We incorporate a different translational motion for the background and include random sub-sequences where the foreground and background undergo the same motion to \cor{simulate} momentarily static objects. As we train our generator with intra-image Fr\'echet loss~($\mathcal{L}_{\mathcal{F}}$), the generated images exhibit strong similarities between the foreground and the background, and even if the object is placed at a different location from the original, we expect it to remain concealed within its surrounding.

\section{Learning to break camouflage}
\label{sec:architecture}
In this section, we train a transformer-based architecture on the synthetic dataset, and demonstrate its effectiveness for breaking the camouflage in real videos, for example, \corrected{MoCA}~\cite{Lamdouar20}. We build on two previous architectures, namely, the motion segmentation from~\cite{Lamdouar21} and the Search Identification Network from~\cite{Fan21}. While the first architecture processes sequences of optical flow, the second takes single RGB images as inputs and treats them separately. 
Our proposed model, shown in Fig.~\ref{fig:data}, takes both the optical flow and RGB frame sequences as input.
The flow is computed with \corrected{RAFT}~\cite{Teed20} from the RGB sequence, and processed with a motion encoder, followed by a transformer encoder.
The RGB sequence is processed with the appearance encoder from~\cite{Fan21}, \modified{pre-trained for framewise camouflage segmentation}. Then 
both streams are \modified{aggregated} as inputs to a \modified{Mask2Former-like}~\cite{cheng22m} transformer decoder and pixel decoder to produce the high resolution segmentation mask from the motion stream. 

\vspace{3pt}
\noindent \textbf{Motion encoder.}
To encode the motion cues, we use a light-weight convNet that takes as input a sequence of optical flows, $\bf{I_{m}}$$=\{I_{m_1},I_{m_2},..,I_{m_t}\} \in \mathcal{R}^{t \times c_0 \times h \times w}$ and outputs motion features:
\begin{align*}
\{f_{m_1},f_{m_2},..,f_{m_t}\} = \Phi_{\text{motion}} (\bf{I_{m}})
\end{align*}
\modified{where} each flow frame is separately embedded.

\vspace{3pt}
\noindent \textbf{Motion transformer encoder.} 
A transformer encoder takes as input the motion features, \modified{concatenated along the sequence}, 
together with \modified{learned spatial and temporal} positional encodings~(\modified{$\text{Pos}_s$, $\text{Pos}_t$}), as indicated in Fig~\ref{fig:data}. 
$\text{Pos}_t$ specifies the frame, and $\text{Pos}_s$ specifies the position in the spatial feature map output of the motion encoder.
The output of the transformer is a set of enriched \modified{motion features.}

\vspace{3pt}

\noindent \textbf{Appearance encoder.} 
Here, we adopt a SINet-v2~\cite{Fan21} architecture, that encodes the RGB sequence, $\bf{I_{a}}$$=\{I_{a_1},I_{a_2},..,I_{a_t}\} \in \mathcal{R}^{t \times c_1 \times h \times w}$ into appearance features:
\vspace{-0.5cm}
\begin{align*}
\{f_{a_1},f_{a_2},..,f_{a_t}\} = \Phi_{\text{app}} (\bf{I_{a}})
\end{align*}
Again, each RGB frame is processed separately by SINet.

\vspace{3pt}
\noindent \textbf{Transformer decoder.}
 A transformer-based decoder takes the output of the motion transformer encoder together with the appearance features and a learnable query for the mask embedding. In a similar manner to \modified{Mask2Former}~\cite{cheng22m}, the query attends to multiple resolutions of the motion features concatenated with the appearance features and produces \modified{a mask embedding for the moving object}. 
 


\vspace{3pt}
\noindent \textbf{Pixel Decoder.}  Similarly to the pixel decoder in Mask2Former,
a light-weight convNet decoder is used with skip-connections to recover high-resolution segmentation masks $\{\hat{m}_1,\hat{m}_2,..,\hat{m}_t\}$ from the motion features and the mask embedding. This is shown as the blue box in Fig.~\ref{fig:data}.

\vspace{2pt}
\noindent \textbf{Training objective.} 
We train the motion segmentation model on our synthetic video sequences using the binary cross-entropy loss $\mathcal{L}_{BCE}$. 

%% file: text/05-experiments.tex

\section{Experiments}
\label{sec:experiment}
In this section, we start by introducing the datasets involved in this paper, followed by the implementation details. 
Our experiments present a thorough analysis of the proposed camouflage scores and demonstrate their effectiveness in our training framework.

\subsection{Datasets}
\begin{figure}[!htb]
\includegraphics[width=0.48\textwidth,height=0.78\textheight]
    {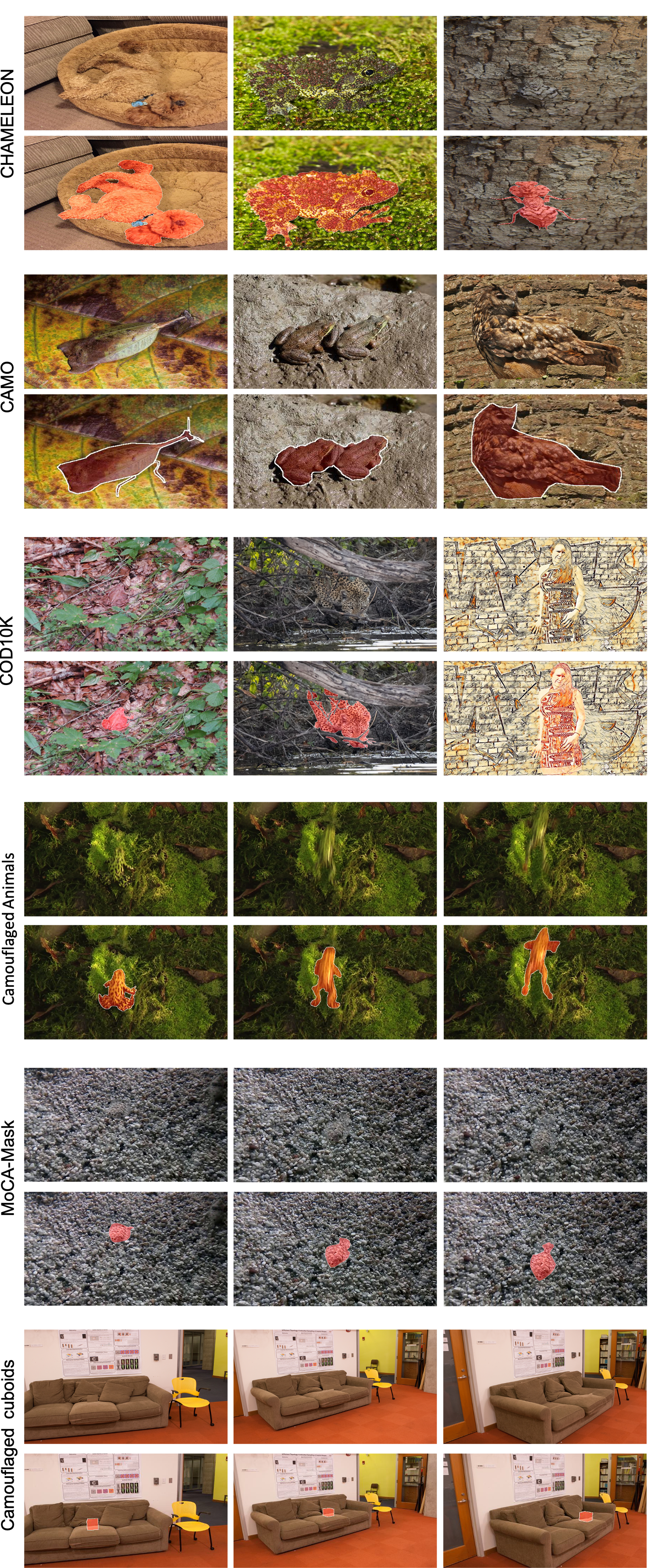}
\caption{\textbf{Randomly sampled examples from the camouflage datasets included in our work}. For the video datasets, Camouflaged Animals and MoCA-Mask, we show an example sequence. For the multiple-view dataset Camouflaged cuboids (bottom), we show example views from a scene with the 4-view texture synthesis method from~\cite{Guo22}.}
\label{fig:datasets}
\vspace{-1.5em}
\end{figure}

Here we describe the publicly available camouflage datasets that we included in our experiments, \modified{as shown in Fig.~\ref{fig:datasets}, as well as} the synthetic camouflage datasets that we generated.

\vspace{2pt}
\noindent \textbf{CHAMELEON~\cite{Skurowski18}} is one of the first camouflage image datasets. It contains 76 images collected using Google image search with the  `camouflaged animal' keyword and include ground truth manual annotations. 

\vspace{2pt}
\noindent \textbf{CAMO~\cite{Le19}.} The Camouflaged Object dataset consists of 2500 images collected from the internet, of which 1250 images (1000 training sub-set and 250 testing sub-set) contain at least one camouflage instance from 8 categories with manual pixelwise annotations provided.

\vspace{2pt}
\noindent \textbf{COD10K~\cite{Fan21}.} COD10K contains 10,000 images collected from photography websites of which 5066 depict camouflaged animals (3040 training sub-set and 2026 testing sub-set), organised into 10 classes, and 78 sub-classes (69 camouflaged). Note that, in our camouflage evaluation experiments we only used the camouflage subset of this dataset.

\vspace{2pt}
\noindent \textbf{Camouflaged Animals~\cite{Bideau16}.} This dataset consists of 9 video sequences of camouflage animals from 6 categories.

\vspace{2pt}
\noindent \textbf{MoCA~\cite{Lamdouar20}} is the first large-scale camouflage video dataset. It contains 141 video sequences (37K frames) totalling 67 animal categories. Recently, other works have curated this dataset by selecting only the videos with more prominent motion (locomotion)~\cite{Yang21a}, and others have provided dense pixel annotation in MoCA-Mask~\cite{Cheng22}. We use the latter version in our experiments. 

\vspace{2pt}
\noindent \textbf{Camouflaged cuboids~\cite{Owens14,Guo22}.} 
This dataset was created for texture synthesis and consists of multiple-view scenes, where cuboids were placed at a predefined location then synthetically camouflaged. In our evaluation, we consider the 4-views generated textures~\cite{Guo22} from 36 scenes as well as the cuboids masks. 

\vspace{2pt}
\noindent \textbf{Synthetic Camouflage Images.}
Using the method described in Sec.~\ref{subsec:gan}, we generate a synthetic camouflage dataset, of $5K$ frames, by discriminating against real camouflage image and annotation masks from COD10K.

\vspace{2pt}
\noindent \textbf{Synthetic Camouflage Videos.}
We generate $1K$ sequences of $30$ frames each, incorporating static sequences using the framework described in Sec.~\ref{subsec:seq}. We split these into $800$ sequences for training and $200$ for testing.

\subsection{Implementation details}
\cor{In our experiments, for a given camouflage example, the kernels for the morphological operations are selected from a range of values [1,10], so that the resulting annotation mask is reduced by 20\% for erosion and extended by 20\% for dilation. This allows the erosion/dilation to be adapted to the size of the camouflaged animal, while always keeping a reasonably large region for addressing the boundary effects.}
\begin{figure*}
  \centering
   \includegraphics[width=0.90\textwidth,height=20em]{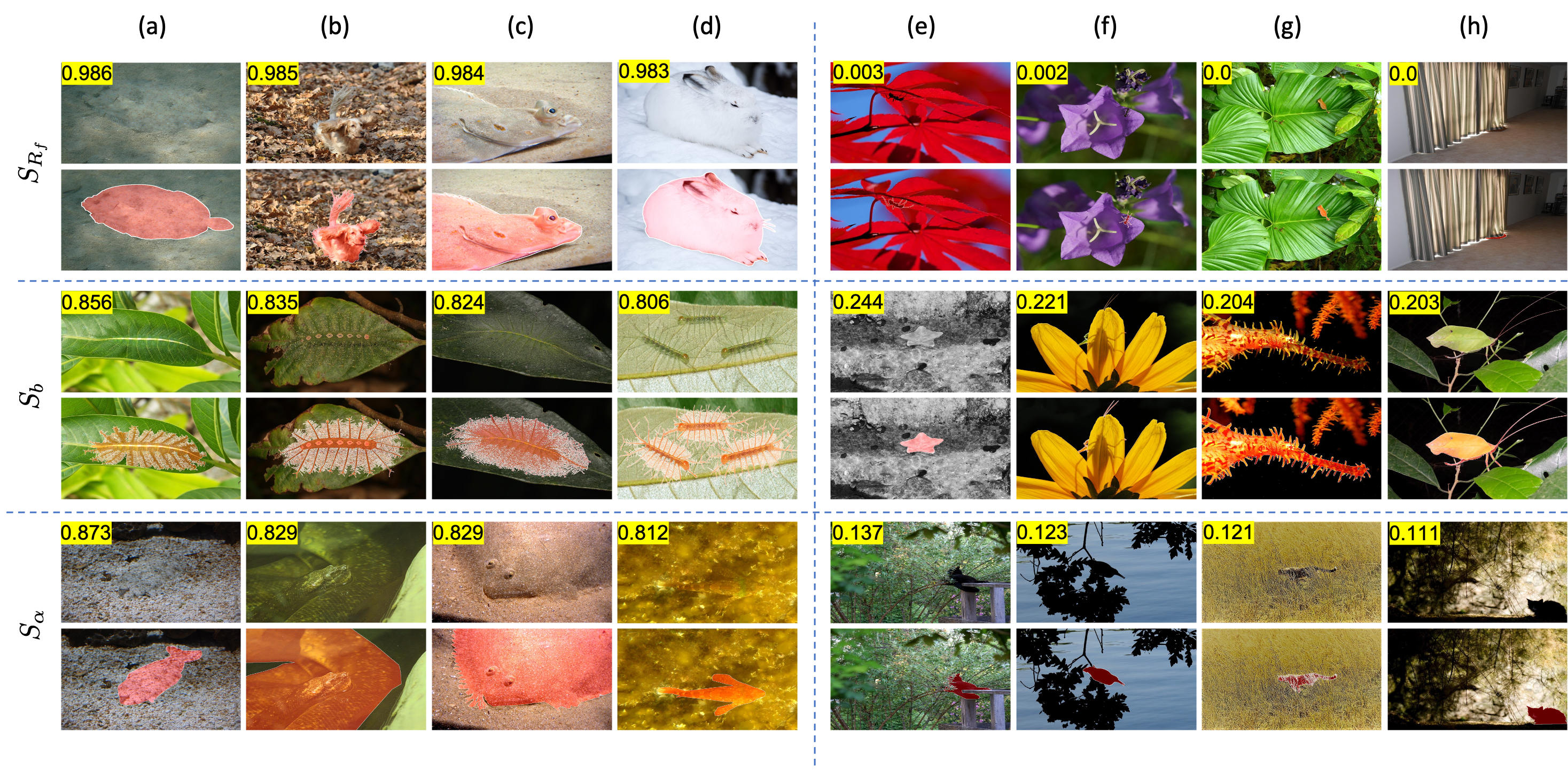}
   \vspace{-10pt}
   \caption{\textbf{Left:} Top-4 scored examples from COD10K for $S_{Rf} (top)$, $S_b (middle)$ and $\score (bottom)$. \textbf{Right:} Lowest scored examples from COD10K. \cor{For each example, we show the obtained score and the corresponding ground truth mask.}}
   \vspace{-0.1cm}
   \label{fig:top-low}
\end{figure*}

\vspace{-4pt}
\noindent For our synthetic camouflage image generation, we adopt a Style-GAN architecture~\cite{Zhang22b} and train it on the camouflage images from COD10K to generate $256 \times 256$ images and masks pairs. 
We use $\beta=0.1$ to weight our intra-image Fr\'echet auxiliary loss. 
For numerical stability, we adopt the Newton Schultz iterations when calculating the matrix square root term in Intra-Image Fr\'echet Inception Distance, with $T=100$ iterations.

\vspace{3pt}
\noindent \cor{When} generating the image sequences, 
we compute optical flows with RAFT~\cite{Teed20}, 
and train our moving camouflaged animal segmentation model. We use batches of size 2 and sequences of size 5. We first train on our synthetically generated dataset with a learning rate of \text{$5\times10^{-4}$} for 500 iterations then fine-tune on the training subset of MoCA-Mask.

\subsection{Results}

\input{tables/metrics}

This section presents qualitative and quantitative results, 
to demonstrate the effectiveness of our score functions and the benefit of including $d^2_{\mathcal{F}}$ in the training loop. 
 \vspace{-0.3cm}

\subsubsection{Evaluation on camouflage effectiveness}

\noindent \textbf{Ranking camouflage images in COD10K.}
We can rank the images based on the effectiveness of the animal's camouflage, 
with our proposed scoring functions. 
In the left part of Fig.~\ref{fig:top-low}, 
we show the four best scored images from the large scale COD10K dataset. 
We can make the following observations:
(i) the best examples with respect to $S_{R_f}$ (top), 
which favours the \textbf{background matching} of the animals without taking into account the boundary region, all exhibit visible boundaries, with the exception of image (a). This is especially noticeable in the rabbit example (d) along the ears and shadow regions;
(ii) the best four examples with respect to the \textbf{boundary score} $S_b$ are all from the caterpillar subclass, mostly the baron type. These insects have thin, 
and transparent boundaries, that makes them the perfect candidates for this score. However, with the exception of (a) and (c), the animals still stands out as they exhibit colors and patterns that are not present within their background;
(iii) the $\score$ \textbf{combining both scores} shows highly effective camouflage examples. In contrast, the right part of Fig.~\ref{fig:top-low} shows the lowest scored images for each approach. 
For $S_{R_f}$, we find examples with low background matching in the first row, 
with two ant examples, {\em i.e.}, (e), (f).
While the boundary score penalises examples with high contrasts between the animal and its background, that results in more visible contours. 
This is also the case for the lowest \score\ score examples.

\vspace{2pt}
\noindent \textbf{Dataset level camouflage comparison.}
We compute the camouflage score for all the camouflage image and video datasets and report the results in Tab.~\ref{tab:camo_scores}. \modified{For the image datasets, the dataset level score is computed as the mean of the image scores; for the video dataset the score is first computed per frame, and per-video average computed, then the dataset level score is the mean of the averages. Our experiments show that} MoCA-Mask~\cite{Lamdouar20,Cheng22} contains the most successful camouflages according to our scoring. While COD10K~\cite{Fan21} subsets are balanced in terms of camouflage effectiveness, we find that, for CAMO~\cite{Le19}, the test subset yields higher scores than the training subset and therefore contains more challenging examples.

\modified{We note that Camouflaged cuboids dataset yields higher $S_{R_f}$ and $S_\alpha$ scores compared to our synthetic datasets. This is due to the fact that the model used in~\cite{Owens14,Guo22} is only trained to produce optimal texture for a predefined region (cuboid) in a particular location of a given scene. However, our model learns to output a {\em new} image with a randomly located camouflaged object of random shape, thus offering more scalability and diversity compared to the cuboids dataset which cannot be used to train a model for breaking real camouflage.}


\vspace{2pt}
\noindent \textbf{Comparison to human-produced rankings.}
We compare our rankings to the human scoring system based on the rating of CHAMELEON~\cite{Skurowski18} and the time-to-find for Camouflaged cuboids~\cite{Guo22}. To search for the optimal $\alpha$ parameter in Equation~(\ref{eq:tri-score}), we select $15$ images from CHAMELEON and compare their \score\ ranking with \modified{ground truth} using kendall-$\tau$ metric~\cite{kendall38}. For $\alpha=0.35$, we obtain kendall-$\tau = 0.51$ on this validation set that we excluded from our test reported in Tab.~\ref{tab:kendalltau}. We can draw the following observations:
(i) we find that the {boundary score} $S_b$ produces more agreement with the \modified{ground truth} ranking than $S_{R_f}$, suggesting that human observers tend to pay more attention to contour visibility than background matching;
(ii) while comparing to the ranking from Camouflaged cuboids, 
we found negative correlations for $S_{R_f}$ and $d_{\mathcal{F}}$, 
we conjecture that this may be due to the nature of the dataset in~\cite{Guo22}, 
{\em i.e.}, all the textured cuboids are synthetically generated with very high background matching; 
(iii) for both experiments, we obtain the strongest agreement with the \score\ score combining both background similarity and boundary blending. 

\vspace{2pt}
\noindent \textbf{\modified{Further analysis of the synthetic datasets.}}
\modified{We adopt the FID}~\cite{Heusel17} \modified{and IS}~\cite{Shaham19} \modified{metrics in Table}~\ref{tab:quality}. \modified{These metrics assess: (i) the similarity between the set of synthetic images output by the generator and the set of real images used to train it (FID); and (ii) the clarity of the object and the diversity of images in terms of classes (IS). While they are not intended for measuring camouflage success, which is the focus of this work, IS could somewhat be (inversely) linked to $S_b$ as the object clarity and boundary visibility are close.}
\noindent \modified{Note that the object clarity (IS) is decreased with $\mathcal{L}_{\mathcal{F}}$ which is the intended purpose of improving camouflage, however, this effect is not maintained in the sequence generation as the animal changes location within its background.}
\input{tables/quality_metrics}
\begin{figure*}
  \centering
   \includegraphics[width=0.90\textwidth, height=9.2em]{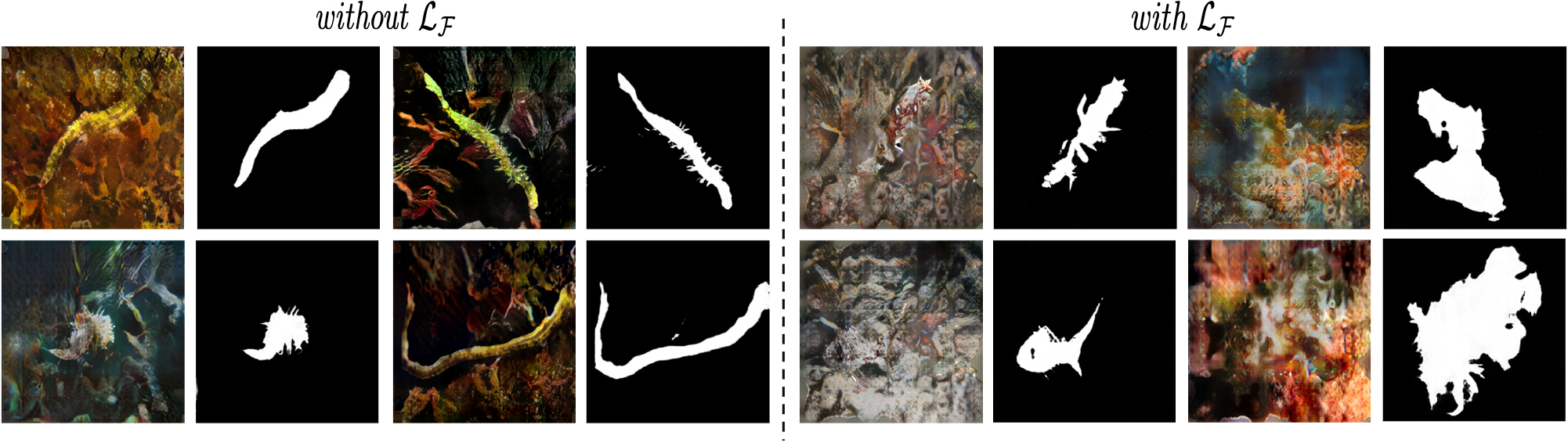}
   \vspace{-5pt}
   \caption{\textbf{Generated camouflage images and masks}: For the examples on the right, the generator was trained with $\mathcal{L}_{\mathcal{F}}$.}
   \label{fig:GAN-samples}
   \vspace{-0.3cm}
\end{figure*}

\begin{figure*}

\begin{minipage}{\textwidth}
\hspace{-0.9cm}
  \begin{minipage}[b]{0.62\textwidth}
\includegraphics[width=0.9\textwidth, height=13em]{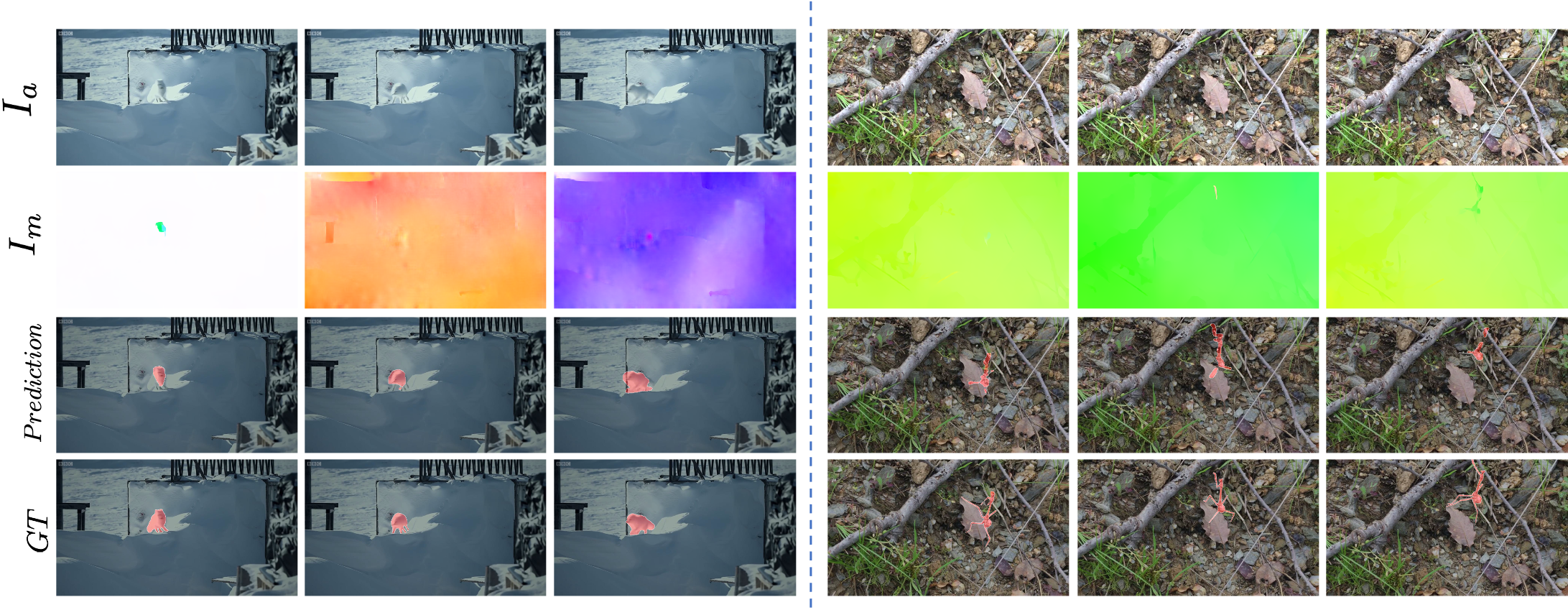}
  \vspace{-0.2cm}
\captionsetup{width=0.8\linewidth}
\centering \captionof{figure}{\textbf{Qualitative results on MoCA-Mask.} \modified{From top to bottom: Appearance sequence $I_a$, flow sequence $I_m$, predicted segmentations and ground truth segmentations.}}
 \vspace{-0.6em} 
   \label{fig:qualitative}

   \end{minipage}
      \hspace{-0.5cm}
   \begin{minipage}[b]{0.45\textwidth}
         
\centering
\setlength{\tabcolsep}{1.5pt}

\begin{tabular}{l|cc|cccc}
\toprule        
 Model  & RGB  & Motion & mIoU$\uparrow$ & \cor{$F\uparrow$} & \cor{$E\uparrow$} & \cor{MAE$\downarrow$} \\ \midrule
SINet~\cite{Fan21} & \checkmark &   &  20.2 & 23.1 & 69.9 & 2.8\\ 
SINet-V2~\cite{Fan21} & \checkmark &   & 18.0 & 20.4 & 64.2 & 3.1\\
\modified{SegMaR}~\cite{jia2022segment} & \checkmark &   &  \modified{ 12.2} & \modified{22.5} & \bf \modified{80.3} & \modified{4.5}\\
\modified{ZoomNet}~\cite{pang2022zoom} & \checkmark &   &  \modified{18.8} &  \modified{28.7} & \modified{70.8} &  \modified{2.5}\\ 
SLTNet~\cite{Cheng22}  & \checkmark &  \checkmark & 27.2 & 31.1 & 75.9 & 2.7\\ 
MG~\cite{Yang21a}  & \ & \checkmark & 12.7 & 16.8 & 56.1 & 6.7\\
Ours-flow  &  & \checkmark  &  17.8 & 21.5 & 60.7 & 3.7 \\ 
Ours &  \checkmark & \checkmark  & \bf 30.8 & \bf 34.3 &  77.0 & \bf 1.8\\   
\bottomrule
\end{tabular}
\vspace{-0.15cm}
\hspace{1cm}
\centering \captionof{table}{\textbf{Results on MoCA-Mask.} \modified{Ours-flow refers to the optical flow only version of our architecture, excluding the appearance stream.}}
\label{tab:moca}
\vspace{-.2cm}
   \end{minipage}
\end{minipage}
\vspace{-.2cm}
\end{figure*}
\vspace{-1pt}
\subsubsection{Model ablations}

\vspace{-4pt}

\noindent \textbf{On the effectiveness of $\mathcal{L}_{\mathcal{F}}$.}
Fig.~\ref{fig:GAN-samples} shows the generated samples from the GAN pipeline, trained with and without the intra-image Fr\'echet Distance.
Our generator is able to produce realistic object masks of complex shapes and thin structures similar to those usually encountered in the camouflage datasets. 
The produced images exhibit rich backgrounds with realistic nature-like elements mimicking rocks and coral structures present in the real camouflage data. 
Adding \cor{the $\mathcal{L}_{\mathcal{F}}$} loss results in generating images with better blending in the background, both qualitatively and quantitatively, as shown by the \score\ computed for both datasets generated with and without the intra-image Fr\'echet Distance in Tab.~\ref{tab:camo_scores}. We refer the reader to the supplementary material for more examples of generated samples. 

\vspace{2pt}
\noindent \textbf{On the architecture design and gain from the synthetic dataset.} \corrected{We present in Tab.~\ref{tab:ablation} an ablation study detailing the gain from the main components of our architecture, as well as the benefit of first training on our synthetic dataset then fine-tuning on MoCA-Mask (S+MM), for model \modified{H}, as opposed to only training on MoCA-Mask (MM), for \modified{G}.}

\vspace{-8pt}
\subsubsection{Results on MoCA}
We train our model on the synthetically generated camouflage images and fine-tune on the training set of MoCA-Mask. Tab.~\ref{tab:moca} presents our camouflage object segmentation results on the test set of MoCA-Mask. \cor{Ours} refers to our proposed method with SINet-V2 as the appearance encoder and Ours-flow refers to the optical flow only version of our architecture, excluding the appearance stream. \cor{Our main model} outperforms RGB and Motion-based methods on MoCA-Mask. Fig.~\ref{fig:qualitative} presents qualitative results of our segmentation model. Note that our model is robust to degraded optical flow and static animals.

\input{tables/moca_sota}
\subsection{Limitations}

While our image generation method encourages background matching through the $\mathcal{L}_{\mathcal{F}}$ loss term minimization, the sequence generating is not guaranteed to maintain object concealment. A possible solution could be to use the proposed \score\ camouflage score to curate the generated sequence dataset and filter out the most visible objects. 

Our proposed camouflage scores use the \modified{ground truth} annotation and analyse the different regions it defines. We found that our scoring system penalises specific cases of camouflage by occlusion, where elements from the background are partially occluding the animal and helping improve its camouflage. For instance, the grass in example (g) from the last row of Fig.~\ref{fig:top-low} is not treated as part of the animal and therefore not considered in our background similarity assessment. One can argue that this is due to the ambiguity in the provided annotations and for such cases, extra amodal annotation should be also considered. \vspace{-1pt}

%% file: tables/metrics.tex
\begin{table*}[!h]
\centering

\sbox0{%
  \begin{minipage}[b]{.55\textwidth}
\footnotesize
\setlength{\tabcolsep}{5pt}
\begin{tabular}{l|ccccc}
\toprule        
 Datasets  & Data type  & $S_{R_f}\uparrow$ & $S_{b}\uparrow$ &
                      \score$\uparrow$ & $d^2_{\mathcal{F}}\downarrow$ \\ \midrule

CHAMELEON~\cite{Skurowski18}  &     Image  &    0.694          &     0.445           &   0.607    & \bf 0.70 \\
CAMO Train~\cite{Le19}                &     Image  &      0.672         &     0.451           &   0.595       &  1.01 \\
CAMO Test~\cite{Le19}                &     Image  &         0.683      &      0.470          &    0.608     &  1.00  \\
COD10K Train~\cite{Fan21}    &     Image   &     0.655          &       0.433         &   0.577       &  0.90
  \\
COD10K Test~\cite{Fan21}        &     Image   &   0.657            &      0.431          &    0.578     &   0.90 \\
Camouflaged Animals~\cite{Bideau16} &     Video &          0.674     &      \bf 0.536          &     0.626    & 1.60   \\
MoCA-Mask Train~\cite{Lamdouar20,Cheng22}  &    Video  &      \bf 0.850         &     0.443           &    \bf 0.707      & 1.14  \\
MoCA-Mask Test~\cite{Lamdouar20,Cheng22}     &    Video  &      0.733          &    0.464            &    0.639     &    2.51 \\ 
\midrule
Camouflaged cuboids\cite{Owens14,Guo22}&    Multi-view   &     \bf 0.894          &     0.433           &    \modified{ \bf 0.733 }     &  6.2\\

Syn. Camouflage w.o. $\mathcal{L}_{\mathcal{F}}$   &     Image   &   0.608 &  0.432  &  0.546 &  1.36\\
Syn. Camouflage w. $\mathcal{L}_{\mathcal{F}}$         &     Image   &     0.679          &     \bf 0.447           &     \modified{ 0.598}      &  \bf 1.13\\
Syn. Camouflage Video  &    Video  &      0.658         &         0.430       &     0.578      &  1.18\\ \bottomrule
\end{tabular}
  \vfill
\vspace{-0.2cm}
\caption{Results of the proposed scores on natural camouflage datasets (top) and synthetically generated camouflage datasets (bottom). We report the mean scores for the single image datasets. For the video and multiple view datasets, we compute the mean per sequence and scene respectively. {\em Syn.\ Camouflage w.\ $\mathcal{L}_{\mathcal{F}}$} refers to the synthetic image dataset that we generated while minimising $\mathcal{L}_{\mathcal{F}}$ and {\em Syn.\ Camouflage Video} its sequence version.
}
\label{tab:camo_scores}
\end{minipage}}
\usebox{0}\hfill
\begin{minipage}[b][\ht0][s]{.43\textwidth}
\footnotesize
\begin{minipage}{\textwidth}
\centering
\begin{tabular}{lcc}
\toprule        
Scores   & CHAMELEON~\cite{Skurowski18} & Camouflaged cuboids\cite{Owens14,Guo22} \\ \midrule
$S_{R_f}$ &  0.01 & -0.07\\
$S_b$  &  0.03 & 0.42\\
\score  &  \bf 0.42 & \bf 0.41\\
$d^2_{\mathcal{F}}$  & 0.10 & -0.17\\
\bottomrule
\end{tabular}
\vspace{-6pt}
\caption{Kendall-$\tau$ metric between rankings produced \modified{via our scores} and human scoring \modified{ground truth}.}
\label{tab:kendalltau}
\end{minipage}

\vfill
\setlength{\tabcolsep}{1pt}
\centering
\resizebox{\textwidth}{!}{
\centering
\begin{tabular}{l|c|ccc|cccc}
\toprule        
 Model  & Training  & Appearance   & Transformer  & Transformer   & mIoU$\uparrow$ & $F\uparrow$ & $E\uparrow$ & MAE$\downarrow$\\ 
  & dataset & Encoder  &  Encoder &  Aggregation  & & & &\\ 
 \midrule
 A & S &  &   &  & 14.5 & 20.4 & 57.3 &  9.4 \\ 
\modified{B} & \modified{MM} &  &   &  & \modified{15.3} & \modified{20.6} & \modified{57.3} & \modified{5.1} \\ 
\modified{C} & S+MM &  &   &  & 16.0 & 21.8 & 59.8 &  4.3 \\ 
\modified{D} & S+MM & \checkmark & &  &  20.5 & 22.6 & 59.8 &  3.8\\ 
\modified{E} & S+MM & \checkmark & \checkmark &  & 23.0 & 23.5 & 57.0 &  \bf 1.6\\ 
\modified{F} & S+MM & \checkmark &  & \checkmark & 22.7 & 23.2 & 58.8 & 3.6 \\ 
\modified{G} & MM & \checkmark & \checkmark  &\checkmark  & 23.4  & 24.7 & 62.0 & 2.4 \\
\modified{H} & S+MM &\checkmark  & \checkmark &\checkmark  & \bf 30.8 & \bf 34.3 & \bf 77.0 & 1.8 \\
\bottomrule
\end{tabular}
}
\vspace{-.3cm}
\caption{Ablation study on the different components of the motion segmentation architecture and training datasets. \cor{We evaluate on MoCA-Mask for all experiments. (S: our synthetic dataset, MM: MoCA-Mask).}}

\label{tab:ablation}
\end{minipage}
\vspace{-0.4cm}
\end{table*}

%% file: tables/quality_metrics.tex
\vspace{-.3cm}
\begin{table}[!htb]
\centering
\small
\setlength{\tabcolsep}{25pt}
\centering
\resizebox{0.49\textwidth}{!}{
\centering
\begin{tabular}{l|c|c}
 dataset                 & FID$\downarrow$  & IS$\uparrow$   \\ \midrule
Syn. Camouflage w.o. $\mathcal{L}_{\mathcal{F}}$ &  \bf 75.01     &     3.49      \\
Syn. Camouflage w. $\mathcal{L}_{\mathcal{F}}$   &   76.34    &   3.13      \\ 
Syn. Camouflage Video   &   86.42    &    \bf 3.85     \\ 
\bottomrule
\end{tabular}
}
\vspace{-.3cm}
\caption{\textbf{Overall quality of our synthetic datasets.}}
\label{tab:quality}
\vspace{-.8cm}
\end{table}

%% file: text/06-conclusion.tex
\section{Conclusion}
We present three score functions for computationally assessing the effectiveness of camouflage in images and videos. By evaluating the similarity with background and the boundary visibility, our combined perceptual score is strongly correlated with human perceptual ranking systems on two different datasets. We demonstrate that training a generative model with our differentiable camouflage function improves the effectiveness of generated camouflage examples and can be used to generate challenging synthetic camouflage datasets to train models to break camouflage. 
\subsection*{Acknowledgement}
\modified{ 
We are grateful to Przemys{\l}aw Skurowski for providing the human perception study data on the CHAMELEON dataset and Andrew Owens, Rui Guo and Oscar de Lima for providing the data for the Camouflaged cuboids dataset. We thank Tengda Han for fruitful discussions. This research is supported by the UK EPSRC funded CDT in Autonomous Intelligent Machines and Systems (AIMS), the EPSRC Programme Grant VisualAI EP/T028572/1, a Schlumberger Studentship, and a Royal Society Research Professorship. WX is supported by the National Key R\&D Program of China (No. 2022ZD0161400).
}

%% file: main.bbl
\begin{thebibliography}{10}\itemsep=-1pt

\bibitem{Bideau16}
Pia Bideau and Erik Learned-Miller.
\newblock It’s moving! a probabilistic model for causal motion segmentation
  in moving camera videos.
\newblock In {\em ECCV}, pages 433--449, 2016.

\bibitem{Bideau18}
Pia Bideau, Rakesh~R Menon, and Erik Learned-Miller.
\newblock Moa-net: self-supervised motion segmentation.
\newblock In {\em Proc. ECCV Workshop}, 2018.

\bibitem{Brox10}
Thomas Brox and Jitendra Malik.
\newblock Object segmentation by long term analysis of point trajectories.
\newblock In {\em European conference on computer vision}, pages 282--295,
  2010.

\bibitem{cheng22m}
Bowen Cheng, Ishan Misra, Alexander~G Schwing, Alexander Kirillov, and Rohit
  Girdhar.
\newblock Masked-attention mask transformer for universal image segmentation.
\newblock In {\em CVPR}, pages 1290--1299, 2022.

\bibitem{Cheng22}
Xuelian Cheng, Huan Xiong, Deng-Ping Fan, Yiran Zhong, Mehrtash Harandi, Tom
  Drummond, and Zongyuan Ge.
\newblock Implicit motion handling for video camouflaged object detection.
\newblock In {\em CVPR}, pages 13864--13873, 2022.

\bibitem{Choudhury22}
Subhabrata Choudhury, Laurynas Karazija, Iro Laina, Andrea Vedaldi, and
  Christian Rupprecht.
\newblock {G}uess {W}hat {M}oves: {U}nsupervised {V}ideo and {I}mage
  {S}egmentation by {A}nticipating {M}otion.
\newblock In {\em BMVC}, 2022.

\bibitem{Doan20}
Khoa~D Doan, Saurav Manchanda, Fengjiao Wang, Sathiya Keerthi, Avradeep
  Bhowmik, and Chandan~K Reddy.
\newblock Image generation via minimizing fr{\'e}chet distance in discriminator
  feature space.
\newblock {\em arXiv preprint arXiv:2003.11774}, 2020.

\bibitem{Efros99}
A. Efros and T. Leung.
\newblock Texture synthesis by non-parametric sampling.
\newblock In {\em Proceedings of the 7th International Conference on Computer
  Vision, Kerkyra, Greece}, pages 1039--1046, September 1999.

\bibitem{Fan21}
Deng-Ping Fan, Ge-Peng Ji, Ming-Ming Cheng, and Ling Shao.
\newblock Concealed object detection.
\newblock {\em IEEE Transactions on Pattern Analysis and Machine Intelligence},
  2021.

\bibitem{frechet57}
Maurice Fr{\'e}chet.
\newblock Sur la distance de deux lois de probabilit{\'e}.
\newblock {\em Comptes Rendus Hebdomadaires des Seances de L'Academie des
  Sciences}, pages 689--692, 1957.

\bibitem{Goodfellow14a}
Ian Goodfellow, Jean Pouget-Abadie, Mehdi Mirza, Bing Xu, David Warde-Farley,
  Sherjil Ozair, Aaron Courville, and Yoshua Bengio.
\newblock Generative adversarial nets.
\newblock In {\em NeurIPS}, pages 2672--2680, 2014.

\bibitem{Guo22}
Rui Guo, Jasmine Collins, Oscar de Lima, and Andrew Owens.
\newblock Ganmouflage: 3d object nondetection with texture fields.
\newblock In {\em arXiv preprint arXiv:2201.07202}, 2022.

\bibitem{Heusel17}
Martin Heusel, Hubert Ramsauer, Thomas Unterthiner, Bernhard Nessler, and Sepp
  Hochreiter.
\newblock Gans trained by a two time-scale update rule converge to a local nash
  equilibrium.
\newblock In {\em Advances in Neural Information Processing Systems}, 2017.

\bibitem{jia2022segment}
Qi Jia, Shuilian Yao, Yu Liu, Xin Fan, Risheng Liu, and Zhongxuan Luo.
\newblock Segment, magnify and reiterate: Detecting camouflaged objects the
  hard way.
\newblock In {\em CVPR}, pages 4713--4722, 2022.

\bibitem{kajiura2021improving}
Nobukatsu Kajiura, Hong Liu, and Shin'ichi Satoh.
\newblock Improving camouflaged object detection with the uncertainty of
  pseudo-edge labels.
\newblock In {\em ACM Multimedia Asia}, pages 1--7. 2021.

\bibitem{kendall38}
Maurice~G Kendall.
\newblock A new measure of rank correlation.
\newblock {\em Biometrika}, 30(1/2):81--93, 1938.

\bibitem{Keuper15}
Margret Keuper, Bjoern Andres, and Thomas Brox.
\newblock Motion trajectory segmentation via minimum cost multicuts.
\newblock In {\em ICCV}, pages 3271--3279, 2015.

\bibitem{Lamdouar21}
Hala Lamdouar, Weidi Xie, and Andrew Zisserman.
\newblock Segmenting invisible moving objects.
\newblock In {\em BMVC}, 2021.

\bibitem{Lamdouar20}
Hala Lamdouar, Charig Yang, Weidi Xie, and Andrew Zisserman.
\newblock Betrayed by motion: Camouflaged object discovery via motion
  segmentation.
\newblock In {\em ACCV}, 2020.

\bibitem{Le19}
Trung-Nghia Le, Tam~V Nguyen, Zhongliang Nie, Minh-Triet Tran, and Akihiro
  Sugimoto.
\newblock Anabranch network for camouflaged object segmentation.
\newblock {\em Computer Vision and Image Understanding}, pages 45--56, 2019.

\bibitem{lezama11}
Jos{\'e} Lezama, Karteek Alahari, Josef Sivic, and Ivan Laptev.
\newblock Track to the future: Spatio-temporal video segmentation with
  long-range motion cues.
\newblock In {\em CVPR}, pages 3369--3376, 2011.

\bibitem{Lin14}
Chiuhsiang~Joe Lin, Chi-Chan Chang, and Yung-Hui Lee.
\newblock Evaluating camouflage design using eye movement data.
\newblock {\em Applied ergonomics}, pages 714--723, 2014.

\bibitem{Mei21}
Haiyang Mei, Ge-Peng Ji, Ziqi Wei, Xin Yang, Xiaopeng Wei, and Deng-Ping Fan.
\newblock Camouflaged object segmentation with distraction mining.
\newblock In {\em CVPR}, pages 8772--8781, 2021.

\bibitem{Ochs11}
Peter Ochs and Thomas Brox.
\newblock Object segmentation in video: a hierarchical variational approach for
  turning point trajectories into dense regions.
\newblock In {\em ICCV}, 2011.

\bibitem{ochs12}
Peter Ochs and Thomas Brox.
\newblock Higher order motion models and spectral clustering.
\newblock In {\em CVPR}, pages 614--621, 2012.

\bibitem{ochs13}
Peter Ochs, Jitendra Malik, and Thomas Brox.
\newblock Segmentation of moving objects by long term video analysis.
\newblock volume~36, pages 1187--1200, 2013.

\bibitem{Owens14}
Andrew Owens, Connelly Barnes, Alex Flint, Hanumant Singh, and William Freeman.
\newblock Camouflaging an object from many viewpoints.
\newblock In {\em CVPR}, pages 2782--2789, 2014.

\bibitem{pang2022zoom}
Youwei Pang, Xiaoqi Zhao, Tian-Zhu Xiang, Lihe Zhang, and Huchuan Lu.
\newblock Zoom in and out: A mixed-scale triplet network for camouflaged object
  detection.
\newblock In {\em CVPR}, pages 2160--2170, 2022.

\bibitem{Poma20}
Xavier~Soria Poma, Edgar Riba, and Angel Sappa.
\newblock Dense extreme inception network: Towards a robust cnn model for edge
  detection.
\newblock In {\em Proceedings of the IEEE Workshop on Applications of Computer
  Vision}, pages 1923--1932, 2020.

\bibitem{Shaham19}
Tamar~Rott Shaham, Tali Dekel, and Tomer Michaeli.
\newblock Singan: Learning a generative model from a single natural image.
\newblock In {\em ICCV}, pages 4570--4580, 2019.

\bibitem{Sivic04a}
Josef Sivic, Frederik Schaffalitzky, and Andrew Zisserman.
\newblock Object level grouping for video shots.
\newblock In {\em Proceedings of the 8th European Conference on Computer
  Vision, Prague, Czech Republic}. Springer-Verlag, 2004.

\bibitem{Skurowski18}
Przemys{\l}aw Skurowski, Hassan Abdulameer, J B{\l}aszczyk, Tomasz Depta, Adam
  Kornacki, and P Kozie{\l}.
\newblock Animal camouflage analysis: Chameleon database.
\newblock {\em Unpublished manuscript}, 2018.
\newblock URL:
  \url{https://www.polsl.pl/rau6/chameleon-database-animal-camouflage-analysis/}.

\bibitem{Suvorov21}
Roman Suvorov, Elizaveta Logacheva, Anton Mashikhin, Anastasia Remizova,
  Arsenii Ashukha, Aleksei Silvestrov, Naejin Kong, Harshith Goka, Kiwoong
  Park, and Victor Lempitsky.
\newblock Resolution-robust large mask inpainting with fourier convolutions.
\newblock {\em arXiv preprint arXiv:2109.07161}, 2021.

\bibitem{szegedy15}
Christian Szegedy, Wei Liu, Yangqing Jia, Pierre Sermanet, Scott Reed, Dragomir
  Anguelov, Dumitru Erhan, Vincent Vanhoucke, and Andrew Rabinovich.
\newblock Going deeper with convolutions.
\newblock In {\em CVPR}, 2015.

\bibitem{szegedy16}
Christian Szegedy, Vincent Vanhoucke, Sergey Ioffe, Jon Shlens, and Zbigniew
  Wojna.
\newblock Rethinking the inception architecture for computer vision.
\newblock In {\em CVPR}, pages 2818--2826, 2016.

\bibitem{Teed20}
Zachary Teed and Jia Deng.
\newblock Raft: Recurrent all-pairs field transforms for optical flow.
\newblock In {\em ECCV}, 2020.

\bibitem{troscianko17}
Jolyon Troscianko, John Skelhorn, and Martin Stevens.
\newblock Quantifying camouflage: how to predict detectability from appearance.
\newblock {\em BMC evolutionary biology}, 17(1):1--13, 2017.

\bibitem{villani09}
C{\'e}dric Villani.
\newblock {\em Optimal transport: old and new}, volume 338.
\newblock Springer, 2009.

\bibitem{xie22}
Junyu Xie, Weidi Xie, and Andrew Zisserman.
\newblock Segmenting moving objects via an object-centric layered
  representation.
\newblock In {\em Advances in Neural Information Processing Systems}, 2022.

\bibitem{Yang21a}
Charig Yang, Hala Lamdouar, Erika Lu, Andrew Zisserman, and Weidi Xie.
\newblock Self-supervised video object segmentation by motion grouping.
\newblock In {\em ICCV}, 2021.

\bibitem{Zhang22b}
Bowen Zhang, Shuyang Gu, Bo Zhang, Jianmin Bao, Dong Chen, Fang Wen, Yong Wang,
  and Baining Guo.
\newblock Styleswin: Transformer-based gan for high-resolution image
  generation.
\newblock In {\em CVPR}, pages 11304--11314, 2022.

\end{thebibliography}
